\documentclass[11pt,a4paper]{article}
\usepackage[hyperref]{emnlp2020}

\usepackage{times}
\usepackage{latexsym}
\usepackage{graphicx}
\usepackage{amsfonts,amssymb}
\usepackage{subfigure}
\usepackage{newunicodechar}
\usepackage{amsmath}
\usepackage{makecell}
\usepackage{multirow}
\usepackage{arydshln}
\usepackage{color,xcolor}
\usepackage{times}
\usepackage{soul}
\usepackage{url}

\usepackage{graphicx}
\usepackage{multirow}
\usepackage{amsmath}
\usepackage{amsthm}
\usepackage{booktabs}
\usepackage{algorithm}
\usepackage{algorithmic,color}

\usepackage{microtype}

\aclfinalcopy 
\aclfinalcopy 

\title{Syntax-aware Data Augmentation for Neural Machine Translation}

\author{Sufeng Duan$^{1,2,3}$, Hai Zhao$^{1,2,3}$\thanks{$^{*}$Corresponding author. This paper was partially supported by National Key Research and Development Program of China (No. 2017YFB0304100) and Key Projects of National Natural Science Foundation of China (No. U1836222 and No. 61733011).}, Dongdong Zhang$^{4}$, Rui Wang$^{5}$  \\
	$^{1}$Department of Computer Science and Engineering, Shanghai Jiao Tong University \\
	$^{2}$Key Laboratory of Shanghai Education Commission for Intelligent Interaction \\ and Cognitive Engineering, Shanghai Jiao Tong University, Shanghai, China\\
	$^{3}$MoE Key Lab of Artificial Intelligence, AI Institute, Shanghai Jiao Tong University \\
$^{4}$Microsoft Research Asia,Beijing, P.R. China \\
 $^{5}$National Institute of Information and Communications Technology (NICT), Kyoto, Japan\\
{\tt 1140339019dsf@sjtu.edu.cn, zhaohai@cs.sjtu.edu.cn}, \\{\tt dongdong.zhang@microsoft.com, wangrui@nict.go.jp} \\}

\date{}

\begin{document}
\maketitle
\begin{abstract}
  \indent Data augmentation is an effective performance enhancement in neural machine translation (NMT) by generating additional bilingual data. 
In this paper, we propose a novel data augmentation enhancement strategy for neural machine translation. Different from existing data augmentation methods which simply choose words with the same probability across different sentences for modification, we set sentence-specific probability for word selection by considering their roles in sentence. We use dependency parse tree of input sentence as an effective clue to determine selecting probability for every words in each sentence. 
Our proposed method is evaluated on WMT14 English-to-German dataset and IWSLT14 German-to-English dataset. 
The result of extensive experiments show our proposed syntax-aware data augmentation method may effectively boost existing sentence-independent methods for significant translation performance improvement.
\end{abstract}
\section{Introduction}
Data augmentation is helpful in deep learning to boost the accuracy and has been used widely in computer vision (CV) \cite{DBLP:conf/nips/KrizhevskySH12}, natural language processing (NLP) \cite{DBLP:conf/acl/IyyerMBD15,sennrich-etal-2016-improving,XieWLLNJN17,edunov-etal-2018-understanding,DBLP:conf/iclr/ArtetxeLAC18,gao-etal-2019-soft,xia-etal-2019-generalized,DBLP:conf/aclnut/LiS19} and other areas. Data augmentation creates additional data by producing variaions of existing data through transformations such as mirroring, random in CV. In NLP tasks like neural machine translation (NMT), 
data augmentation is used 
to improve the performance by generating additional training samples \cite{gao-etal-2019-soft,xia-etal-2019-generalized,DBLP:conf/iclr/ArtetxeLAC18} or enhance the model robustness by adding explicit noise \cite{DBLP:conf/acl/IyyerMBD15,XieWLLNJN17}.

To perform unsupervised or semi-supervised NMT training with only monolingual data, back-translation \cite{sennrich-etal-2016-improving,edunov-etal-2018-understanding}, which is one kind of data augmentation method at \textit{sentence level}, has been wisely used to generate bilingual data. However, back-translation is not a simple method which requires a full NMT system to translate the target sentence into source. Collecting and cleaning huge amount of monolingual data also require substantial efforts. Some sentences generated by back-translation are not native and are harmful for training , especially for low-resource languages \cite{DBLP:journals/corr/abs-1911-01986}. 

Compared with back-translation, existing data augmentation methods at \textit{word level} include randomly swapping words \cite{DBLP:conf/iclr/ArtetxeLAC18}, dropping words \cite{DBLP:conf/acl/IyyerMBD15}, replacing one word with another \cite{XieWLLNJN17} is simpler and more efficient. These methods focus on changing words in one sentence instead of generating a new entire sentence, though are capable of still creating 
diverse variants of the original sentence.

Existing data augmentation methods for NMT have been shown generally effective. In the meantime, however, they suffer from specific shortcomings, too. The existing methods select word randomly with the same possibility without considering roles of words in one sentence, which may easily lead to generating flawed sentence from replacing or dropping salient words in the original sentence. 
For example, if an important verb in sentence is selected and replaced by other word, then the meaning of the entire original sentence will be totally modified. In addition, if a word is replaced by an improper one or dropped in an improper way, then syntactic structure of the original sentence may be compromised. 
So that the original alignment between source and target sentences may be improperly influenced and the performance of such data augmentation will be limited. 

To alleviate the obvious drawbacks of existing data augmentation methods, 
in this paper, we propose a novel improved data augmentation strategy for NMT. Different from the existing ones 
which generate monolingual data by replacing words with the same possibility, we heuristically select words for revising by 
considering their roles in one sentence. 
In detail, we adopt syntactic dependency parse tree over a sentence as  heuristic clues.
In one dependency parse tree, words have different positions and depths which can be viewed as importance and necessity to the sentence. Generally speaking, the more close to the root in the tree, the more important a word is. In practice, We reduce the possibility of word to be selected if this word is closer to the root of tree compared to other words in the sentence. Thus We actually prefer to select words which are far from the root which are supposed less important. 
Our syntax-aware enhancement method may conveniently incorporate with standard data augmentation operations at word level such as swapping, dropping and replacing.

We evaluate our method on WMT14 English-to-German and IWSLT14 German-to-English datasets. The result of experiments show that our proposed strategy may effectively 
help baseline data augmentation methods for significant performance improvement.

\section{Related Works}
\subsection{Neural Machine Translation} 
NMT models are based on a sequence-to-sequence (seq2seq) architecture, which uses an encoder to create a vector of the source sentence and a decoder to generate the target as a sequence of target words, along with an attention \cite{kalchbrenner2013recurrent,sutskever2014sequence,bahdanau2014neural}. A series of seq2seq NMT model, such as RNN model \cite{sutskever2014sequence,bahdanau2014neural}, CNN \cite{gehring2017convolutional} model and the Transformer \cite{vaswani2017attention}, have been proposed and received better and better performance.

For the Transformer being the state-of-the-art NMT model, in this work, we take it as our baseline model. The Transformer is a fully attention-based NMT model empowered by self-attention networks which is proposed in \cite{DBLP:conf/iclr/LinFSYXZB17}. Encoder of the Transformer consists of one self-attention layer and a position-wise feed-forward layer. Decoder of the Transformer contains one self-attention layer, one encoder-decoder attention layer and one position-wise feed-forward layer. The Transformer uses residual connections around the sublayers and then followed by a layer normalization layer.
Scaled dot-product attention \cite{vaswani2017attention} is the key component in the Transformer. \cite{vaswani2017attention} propose multi-head attention which is used in the Transformer to generate representation of sentence by dividing queries, keys and values to different heads and get information from different subspaces. 

\subsection{Dependency Parsing}

As a fundamental NLP task, (syntactic) dependency parsing aims to predict the existence and type of linguistic dependency relations between words in a sentence 
\cite{DBLP:conf/conll/LiHZZ18,DBLP:conf/emnlp/LiHCZZLLS18,DBLP:conf/acl/ZhaoHLB18}. 

Dependency parsers may be roughly put into two categories in terms of searching strategies over parsing trees, graph-based and transition-based \cite{DBLP:conf/coling/LiCHZ18}. With the development of neural network applied to dependency parsing, there comes continuous progress for better parsing performance \cite{DBLP:conf/conll/WangZZ17,DBLP:conf/conll/LiHZZ18}. 
Zhang et al. \cite{DBLP:conf/acl/ZhangZQ16} propose a neural probabilistic parsing model which explores up to third-order graph-based parsing with maximum likelihood training criteria. Li et al. \cite{DBLP:conf/aaai/LiZJZ18} propose a full character-level neural dependency parser together with a released character-level dependency treebank for Chinese. Dependency parsing is shown to be more effective than non-neural parser. Wu et al. \cite{DBLP:conf/conll/WuZT18} propose a system  for multilingual universal dependency parsing from raw text. Li et al.  \cite{DBLP:conf/pricai/LiCZ19} propose a tree encoder and integrate pre-trained language model features for a better representation of partially built dependency subtrees and thus enhances the model. 

\begin{figure*}[h]
    \centering
    \includegraphics[scale=0.265]{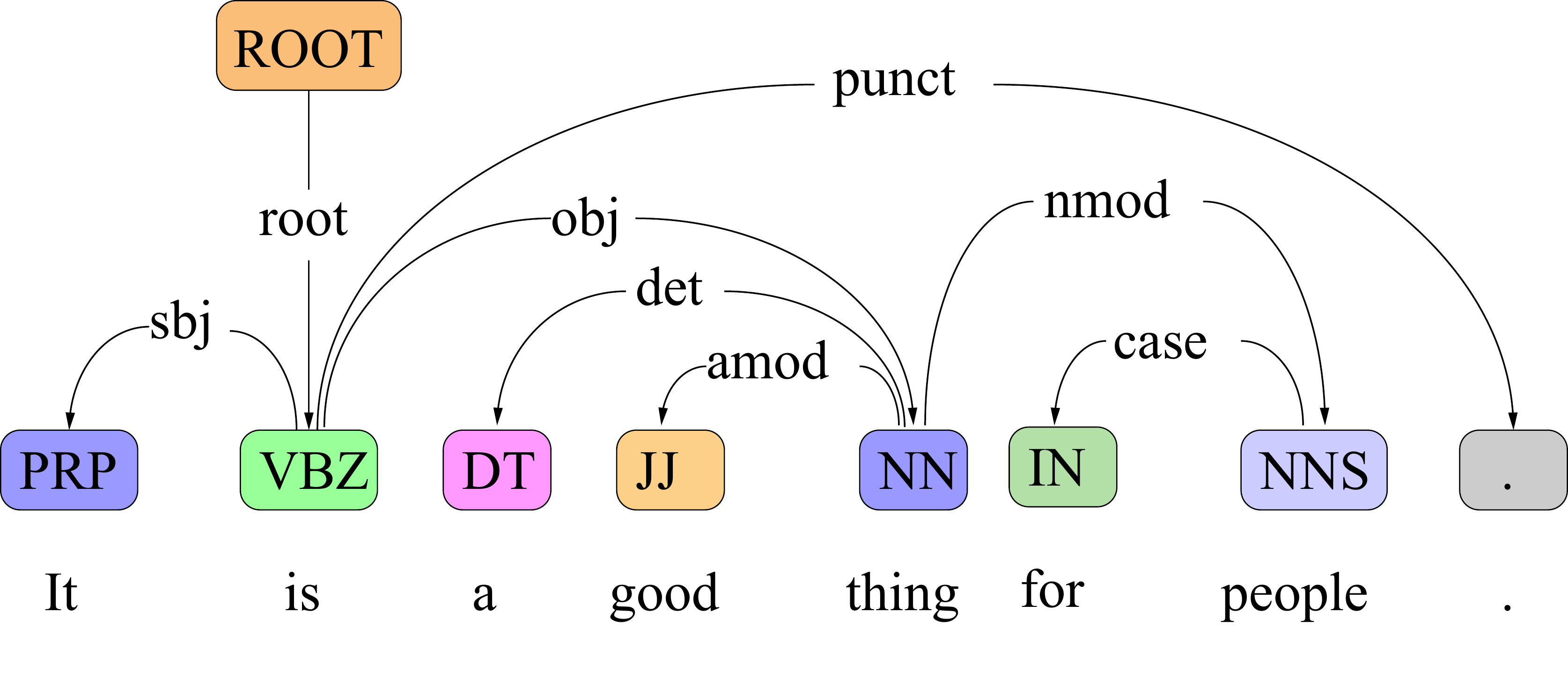}
    \caption{An example of dependency parsing tree.}
    \label{fig:dependency}
\end{figure*}

Figure \ref{fig:dependency} illustrates dependency parse tree for the sentence \textit{It is a good thing for people}. The tree has only one root and every word in this sentence has one and only one parent. The label between one word and its parent reflects the relationship between them. 

Dependency parse tree may be viewed as one pre-trained information for NMT, which has been incorporated into NMT for better translation. 
Eriguchi et al. \cite{eriguchi2016tree} propose a tree-to-sequence model with a tree-based encoder which encodes the phrase structure of a sentence as vectors. Aharoni and Goldberg \cite{aharoni2017towards} design a sequence-to-tree model which translates source sentence to a linearized constituency tree. Method this paper proposed can also be viewed as another way to incorporate dependency parsing information into NMT model.

\subsection{Data Augmentation}\label{sec:da}
Data augmentation is a training enhancement paradigm that has been broadly used in CV \cite{DBLP:conf/nips/KrizhevskySH12,DBLP:conf/icml/WanZZLF13,DBLP:conf/nips/TranPCP017,DBLP:journals/access/LemleyBC17} and NLP \cite{DBLP:conf/acl/IyyerMBD15,sennrich-etal-2016-improving,XieWLLNJN17,edunov-etal-2018-understanding,DBLP:conf/iclr/ArtetxeLAC18,gao-etal-2019-soft,xia-etal-2019-generalized,DBLP:conf/aclnut/LiS19}. 

In NMT, data augmentation is used for either better model robustness by generating more noisy data or better performance by generating more helpful training samples. 
As a kind of data augmentation approach at sentence level,  
back translation has been effectively adopted by unsupervised NMT \cite{sennrich-etal-2016-improving,edunov-etal-2018-understanding}, 
\cite{DBLP:conf/acl/IyyerMBD15,XieWLLNJN17,DBLP:conf/iclr/ArtetxeLAC18,DBLP:conf/iclr/LampleCDR18}, in which 
data augmentation operation is essentially used to facilitate unsupervised NMT by generating data from monolingual corpora. 
In addition, back translation may also improve the performance of supervised NMT. 

As for data augmentation at word level, there exist multiple approaches.

Artetxe et al. \shortcite{DBLP:conf/iclr/ArtetxeLAC18} swap words randomly with nearby words within a window size. Iyyer et al. \shortcite{DBLP:conf/acl/IyyerMBD15} randomly drop some words of one sentence. Xie et al. \shortcite{XieWLLNJN17} propose two methods to add noisy to sentence, replacing words with a placeholder word randomly and replacing words with other words having similar frequency distribution over the vocabulary. 

For the Transformer based NMT, the method in \cite{DBLP:conf/iclr/ArtetxeLAC18} can also be viewed as a special noisy-adding method for position encoding which does not apply impact over the word directly. Positions of words in the Transformer is considered as a feature for input sequence and relative positions of pairs of words are learned by scaled dot-product attention. It makes this method different from replacing words for the position to add noisy.

\section{Method}
For NMT, data augmentation is used for either better model robustness by generating more noisy data or better performance by generating more helpful training samples. In this work, we focus on improving word-level data augmentation approaches for a purpose of model robustness enhancement. 
Such data augmentation methods for NMT usually 

select some words with a fixed sampling probability from source sentences, then apply an altering operation over the selected word to
generate a variant of the original sentence. The new source sentences will be still aligned to the same target sentence as a training sample for NMT training.

As the related work described in Section \ref{sec:da}, we will work on three representative data augmentation operations from which as follows,
\begin{itemize}
\item \textbf{Blanking} \cite{XieWLLNJN17}. Words in sentence will be randomly replaced with a special placeholder token \textit{BLANK}.
\item \textbf{Dropout} \cite{DBLP:conf/acl/IyyerMBD15}. Words in sentence will be randomly dropped by simply setting the respective word embedding as zero vector. 
\item \textbf{Replacement} \cite{XieWLLNJN17}. Words in sentence will be randomly selected and replaced with one word which has a similar unigram word frequency over dataset.
\end{itemize}

We give an example sentence for data augmentation here. Suppose we have one sentence \textit{We shall fight on the beaches.} and we want to get  extra sentences by data augmentation method at word level. Table \ref{table:example} shows sentences generated by different methods.

\begin{table*}[t]
\centering
\begin{tabular}{l|ccccccc}
\hline
Original&We & shall & fight & on & the & beaches & .\\
\hline
Select or Not&no&no&no&yes&no&yes&no\\
\hline
\textit{Blanking}&  We & shall & fight & \textcolor{blue}{\textit{BLANK}} & the &  \textcolor{blue}{\textit{BLANK}}  &.\\
\textit{Dropout}& We & shall & fight & \textcolor{blue}{\textit{ }} & the &  \textcolor{blue}{\textit{ }}  &.\\
\textit{Replacement} & We & shall & fight & \textcolor{green}{\textit{with}} & the &  \textcolor{green}{\textit{sandy}}  &.\\
\hline
\end{tabular}
\caption{An example of generating sentence by data augmentation methods.}
\label{table:example}
\end{table*}

\subsection{Basic Idea}

Our proposed syntax-aware method is an enhancement over standard data augmentation at word level, which applies word-level operations such as swapping, dropping and replacing over every selected words. Straightforward word selecting in existing methods takes a completely sentence-independent (i.e., context-free) strategy for every words in every sentences, which does not show efficient enough. Thus we propose using syntactic clues to guide such a word selection, which will result in a sentence-specific (context-dependent) strategy.

It is well known that the meaning of a sentence is actually determined by only a few of important words. Thus modifying those syntactically and semantically more important words may alter the sentence more radically. For the purpose of robustness-oriented data augmentation, it is to improve the translation robustness by intentionally introducing moderate noisy data. However, too much radically altered sentences may become a truly harmful noise which eventually hurts the NMT training.

Thus, 
for the purpose of improving the model robustness,
effective data augmentation methods should
1) choose large enough number of words for word-level operations, and 2)  abstain from choosing too important words. To meet such requirements, we need to find a heuristic clue to measure how much important a word is, and then determine a selecting probability to alter the corresponding word for data augmentation. 

\subsection{Dependency Tree Depth as Clue} 

Considering that the root of one dependency parsing tree figures out the most salient word in a sentence (which is usually a key \textit{verb} in linguistics), and parent is always more essential than its children according to dependency grammar, we use the distance between word and the root, or the depth of word in the parse tree as our initial clue to measure the importance of a word. Therefore, it makes leaf node of dependency parsing tree more likely to be selected during data augmentation process. Figure \ref{fig:dependency} shows an example of word's depth in dependency tree.

In this work, we only use depth of word in a dependency parse tree and exclude other information delivered by the dependency tree such as dependency relationship or label among any word pair, whose reason is twofold. 1) As we adopt the Transformer as our NMT model baseline, it has shown capacity that learns the relationship between any word pair through its powerful self-attention mechanism. 2) We have a purpose to seek a clear enough heuristic clue indicating the importance of a word in its sentence, which needs to measure the relationship between the word and the entire sentence, rather than the relationship of a word pair offered by the dependency tree.

Different from existing data augmentation methods, which commonly adopt word frequency as a straightforward clue to select words for revising, our adopted syntax-aware clue may enjoy an obvious advantage that word selecting is specific to its sentence, rather than the case that all the same words in different sentences have to be selected simply according its frequency in the whole dataset.

\begin{figure}[h]
    \centering
    \includegraphics[scale=0.4]{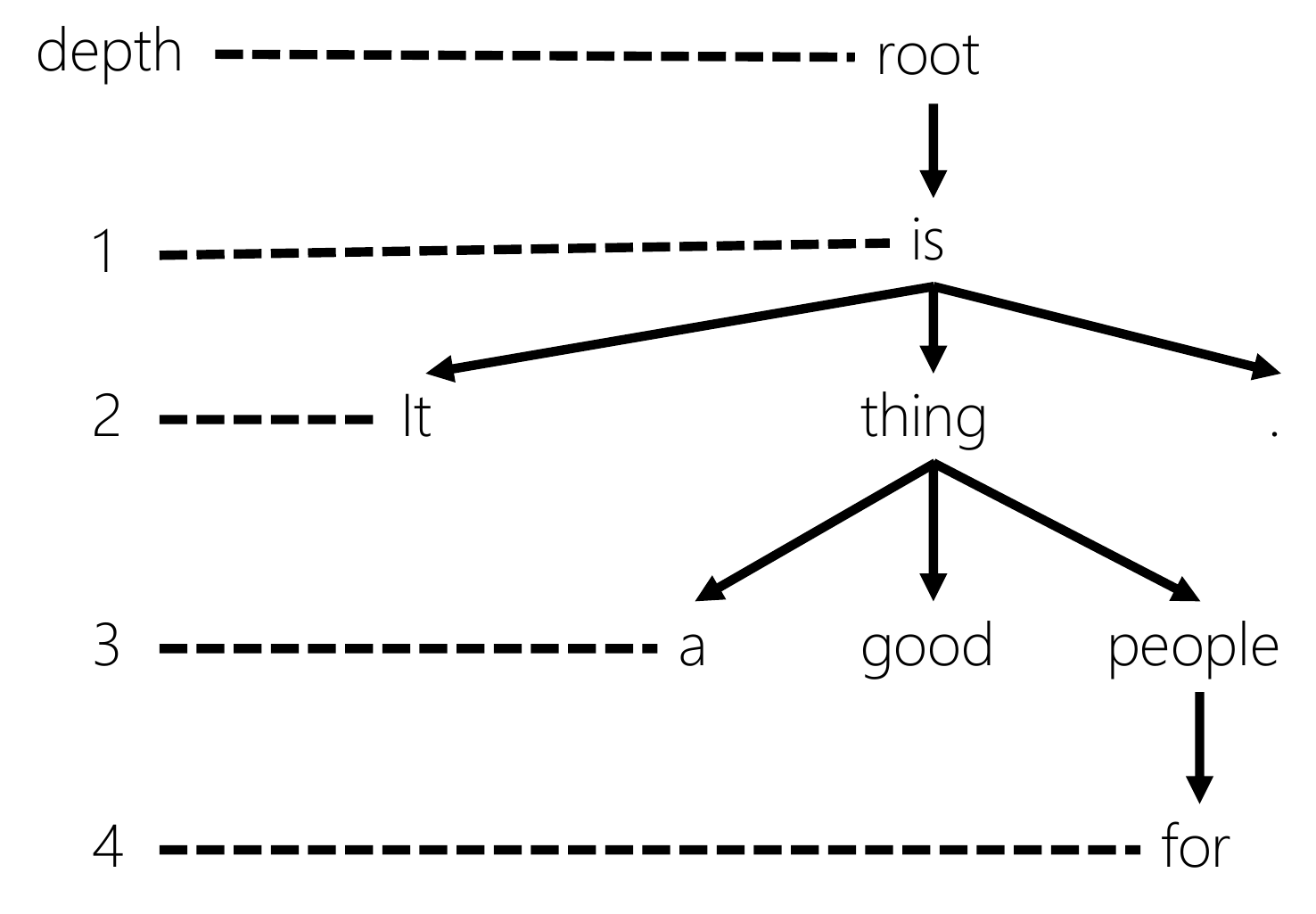}
    \caption{Depths of words in dependency tree.}
    \label{fig:deps}
\end{figure}

\subsection{Probability of Word Selection}

Taking the tree depth as an initial heuristic clue,
we assume a reasonable word selection probability design should satisfy the following conditions, 
\begin{itemize}
 \item The resulting probabilities of words with different tree depths should have diversity for sufficient distinguishability. 
 \item Words in different sized (especially long) sentences should have an appropriate chance to be modified. 
 \end{itemize}

 Given a sentence $s$=${w_1, w_2, ..., w_n}$ of length $n$,

 we first calculate a probability $q_i$ based on $d_i$, the depth of word $w_i$, by
 \begin{equation}
     \begin{aligned}
     q_i  =1-\frac{1}{2^{d_i-1}},
     \end{aligned}
 \end{equation}
 where $d_i$ is the depth of word $w_i$. According to Eq. (1), $q_i$=0 when $w_i$ is right the root node ($d_i$=1). 
 
\begin{table*}[t]
\centering
\begin{tabular}{lcccccccc}
\hline
&It & is & a & good & thing & for & people &.\\
\hline
Depth& 2 & 1 & 3 & 3 & 2 & 4 & 3 & 2 \\
$q_i$&0.5&0&0.75&0.75&0.5&0.875&0.75&0.5\\
$p_i$&0.112&0.068&0.144&0.144&0.112&0.163&0.144&0.112\\
$p^f_i$&0.089&0.054&0.115&0.115&0.089&0.131&&0.089\\
selected&no&no&no&yes&no&yes&no&no\\
\hline
$Blanking$&\textcolor{blue}{\textit{BLANK}}&is&a&good&thing&&\textcolor{blue}{\textit{BLANK}}&.\\
Syntax-aware $Blanking$ &It&is&a&\textcolor{blue}{\textit{BLANK}}&thing&\textcolor{blue}{\textit{BLANK}}&people&.\\
\hline
\end{tabular}
\caption{Computing the probability of selecting words. The last two lines are the result of the original data augmentation $Blanking$ and ours. $Blanking$ is to replace word with a special placeholder \textcolor{blue}{\textit{BLANK}}.}
\label{shows}
\end{table*}

 Second, we let all $Q$=\{$q_i$\} pass a processing of softmax layer to get an adjusted probability distribution $P$ for $s$,

 \begin{equation}
 \begin{aligned}
 P(s) &= {\rm Softmax}(Q(s)) \\
 &= \{p_{1}, p_{2}, ..., p_{n}\},
\end{aligned}
\label{eqsofs}
\end{equation}

At last, to introduce sentence length compensation so that words in longer sentence have proportionable possibility to be selected, we let the final word selection probability be

\begin{equation}
    \begin{aligned}
        p^f_i=\alpha p_{i}n, 
    \end{aligned}
    \label{eqps}
\end{equation}
where $p^f_i$ is the final possibility for selecting word $w_i$ and $\alpha$ is a hyperparameter to control the magnitude of possibility changing, which is set to $\alpha$ to 0.1 in this work. 

Using the dependency tree example in Figure \ref{fig:deps}, we demonstrate a procedure of computing the probabilities for word selecting as shown in Table \ref{shows}. 
According to Eq. (1) and Eq. (\ref{eqps}), we get the initial and final probabilities of word. The larger a word depth is, the more likely the word be selected. From the fifth line in Table \ref{shows}, we can see 
that with the smallest depth, possibility of \textit{thing} is lower than other nodes such as leaf node \textit{for}.

\section{Experiments}

In this paper, data augmentation will only process source data from the training data. The translation quality is evaluated by case-sensitive sacreBLEU score.
\subsection{Datasets}
Two translation tasks, IWSLT14 German-to-English (De-En) and WMT14 English-to-German (En-De), are used for our evaluation.

\textbf{IWSLT14 German-English} IWSLT14 De-En dataset contains 153K training sentence pairs. We use 7K data from the training set as validation set and use the combination of dev2010, dev2012, tst2010, tst2011 and tst2012 as test set with 7K sentences which are preprocessed by script\footnote{https://github.com/eske/seq2seq/blob/master/config/IWSL-T14/prepare-mixer.sh}. BPE algorithm is used to process words into subwords, and number of subword tokens in the shared vocabulary is 31K. 

\textbf{WMT14 English-German} We use the WMT14 En-De dataset from Stanford\footnote{https://nlp.stanford.edu/projects/nmt/} with 4.5M sentence pairs for training. We use the combination of newstest2012 and newstest2013 as validation set and newstest2014 as test set. The  sentences longer than 80 are removed from the training dataset. Dataset is segmented by BPE so that number of subwords in the shared vocabulary is 32K.

Stanford Parser\footnote{https://nlp.stanford.edu/software/lex-parser.html} is used to process German corpus to get  dependency tree and POS tags. For English corpus, we use Stanford Dependency Parser\footnote{https://nlp.stanford.edu/software/nndep.html} to get  dependency tree.

\begin{table}[htb]
\caption{Hyperparameters for our experiments. FF is short for feed-forward layer. The number of heads is based on the dimension for word and feature. }
\newcommand{\tabincell}[2]{\begin{tabular}{@{}#1@{}}#2\end{tabular}}
\begin{center}
\begin{tabular}{lccccc}
\toprule
\multirow{1}{*}{\vspace{-2mm}Parameter} & \multicolumn{1}{c}{DE-EN}&\multicolumn{1}{c}{EN-DE} &  \\

\hline
\tabincell{c}{Layers}&6&6 \\
\tabincell{c}{Dimension } &512&512 \\ 

\tabincell{c}{Head} &\tabincell{c}{4}&\tabincell{c}{8} \\
\tabincell{c}{FF} &1024&2048 \\
\tabincell{c}{Dropout} &0.3&0.1 \\
\bottomrule
\end{tabular}
\end{center}

\label{hyperparameters}
\end{table}

\subsection{Hyperparameters}
The hyperparameters for our experiments are shown in Table \ref{hyperparameters}. For De-En, we follow the setting of Transformer-small. For En-De, we follow the setting of Transformer-base. The input embedding size of our model is from summing up the dimensions of word embeddings and syntactic features. 

\subsection{Training}
All our models are trained on one CPU (Intel i7-5960X) and one nVidia 1080Ti GPU. The implementation of model is based on fairseq-0.6.2. We choose Adam optimizer with $\beta_1=0.9$, $\beta_2=0.98$, $\epsilon=10^{-9}$ and the learning rate setting strategy, which are all the same as \cite{vaswani2017attention},

$lr$ = $d^{-0.5}$ $\cdot$ min($step^{-0.5}$, $step$ $\cdot$ $warmup_{step}^{-1.5}$)
\noindent where $d$ is the dimension of embeddings, $step$ is the step number of training and $warmup_{step}$ is the step number of warmup. When the number of step is smaller than the step of warmup, the learning rate increases linearly and then decreases. 

We use beam search decoder for De-En task with beam width 6. For En-De, following \cite{vaswani2017attention}, the width for beam search is $6$ and the length penalty $\alpha$ is 0.2. The batch size is 1024 for De-En and 4096 for En-De. We evaluate the translation results by using sacreBLEU.
                        
\begin{table}[htb]
\newcommand{\tabincell}[2]{\begin{tabular}{@{}#1@{}}#2\end{tabular}}

\begin{center}
\begin{tabular}{lcccc}
\toprule
\multirow{2}{*}{\vspace{-2mm}Model} & &\multicolumn{2}{c}{sacreBLEU} &  \\
\cmidrule{3-4}
& & DE-EN&EN-DE  \\
\hline
Transformer (small) &&  36.5 & - \\
Transformer (base) && - & 26.8 (27.1) \\
\hline
Blanking &&36.1&26.9 (27.2)\\
Dropout&&36.3&26.8 (27.1)\\
Replacement&&36.0&26.9 (27.2)\\
\hline
\tabincell{c}{Our Method (Blanking) }&& 36.8 & 27.6 (27.9)\\
\tabincell{c}{Our Method (Dropout)}&&36.4&27.0 (27.3) \\
\tabincell{c}{Our Method (Replacement)} && 36.2 & 26.7 (27.0)\\
\bottomrule

\end{tabular}
\end{center}
\caption{BLEU scores on IWSLT14 De-En and WMT14 En-De. The baselines for De-En task and En-De task are the Transformer-small and the Transformer-base, respectively. We use multiBLEU for IWSLT14 De-En, and we use both sacreBLEU and multiBLEU (in parentheses) for WMT14 En-De}
\label{bleu1}
\end{table}

\section{Results}
\label{results}
Our baselines for WMT14 En-De and IWSLT14 De-En are Transformer-base and Transformer-small. Table \ref{bleu1}  compares our data augmentation methods with the original data augmentation methods, showing that our method enhances all the original data augmentation in De-En and En-De tasks and outperforms all baselines. 

On WMT14 En-De translation task, our method on \textit{Blanking} gets the highest BLEU score 27.6, and achieves more than 1.0 BLEU score improvement over the Transformer-base and 0.5 BLEU score improvement over the original \textit{Blanking}. The original \textit{Replacement} method gets the lowest BLEU score in WMT14 En-De with only 24.8 (i.e., actually it hurts the performance over baseline) while our method help it boost to 26.6 which is even higher than baseline.

Compared our method with the baseline data augmentation methods which fail to receive performance improvement, our method helps all the three data augmentation methods outperform baseline and the original data augmentation methods. The experiment result shows that our method can improve the performance of data augmentation methods on both small scale IWSLT14 De-En and WMT14 En-De dataset.

\begin{table}[htb]
\newcommand{\tabincell}[2]{\begin{tabular}{@{}#1@{}}#2\end{tabular}}

\begin{center}
\begin{tabular}{lccc}
\toprule
\multirow{1}{*}{\vspace{-2mm}Model} & &\multicolumn{2}{c}{sacreBLEU} \\
\hline
Transformer (base) && 26.5 \\
\hline
Blanking &&27.1\\
\tabincell{c}{Our method (Before) }&&27.6\\
\tabincell{c}{Our method (During) }&&27.6\\
\bottomrule

\end{tabular}
\end{center}
\caption{BLEU scores on WMT14.}
\label{bleu2}
\end{table}

To valid the performance of our method in different stages of training, we do data augmentation \textit{Blanking} with our method before training and during training respectively on WMT14 En-De. Table \ref{bleu2} shows that there is no significant performance difference, which means our method can be effectively applied in different stages of training.

\begin{figure}
    \centering
    \includegraphics[scale=0.55]{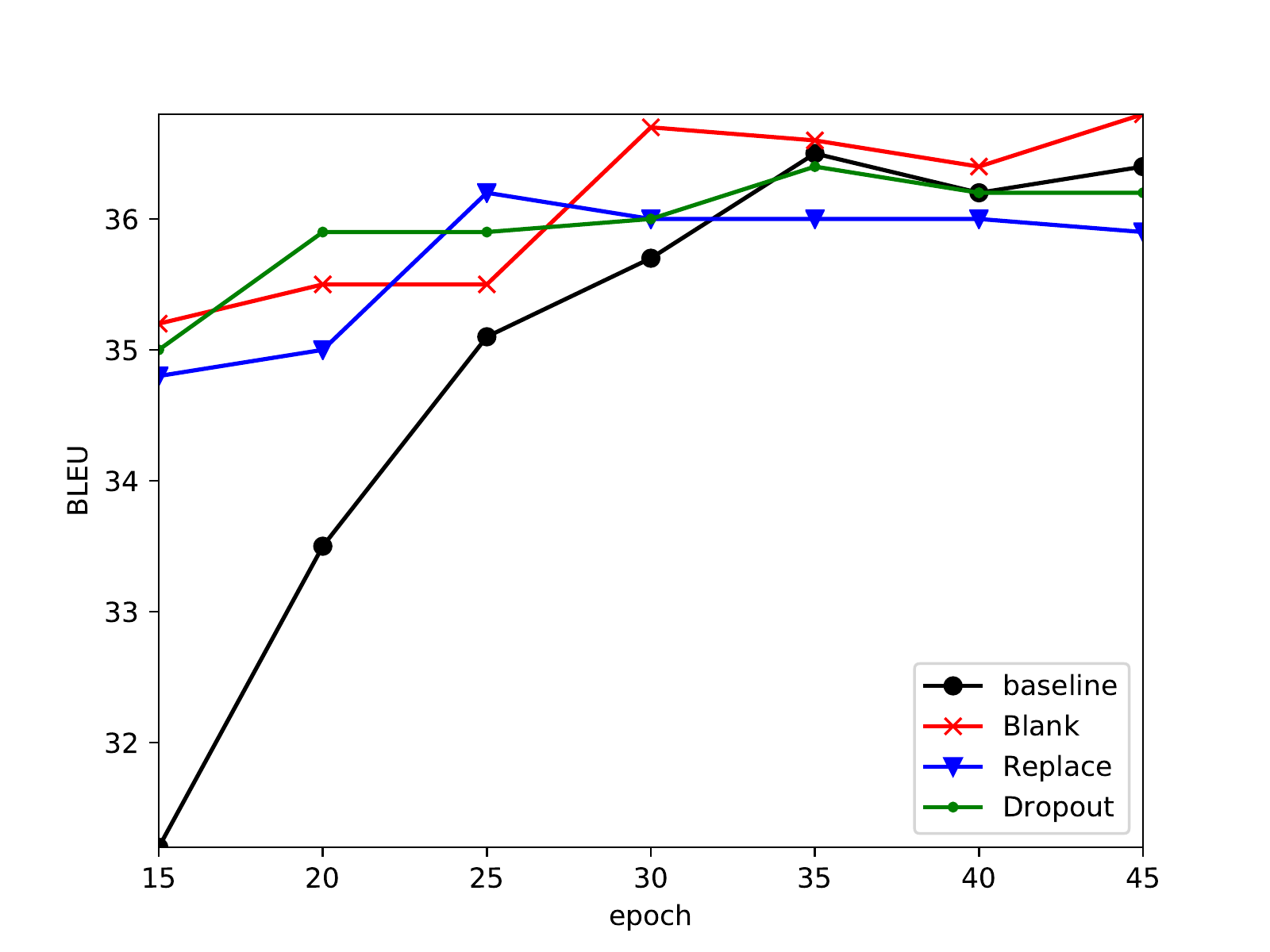}
    \caption{BLEU score in different epoch.}
    \label{fig:convergence}
\end{figure}
To valid the performance of convergence of our method, we calculate the BLEU score with our method in different epoch. Figure \ref{fig:convergence} shows that all our methods can converge faster compared with our baseline. Specially, we can see our methods can achieve a better performance at beginning, which means our method can significantly improve the speed of convergence.

\section{Discussion}
As mentioned, our proposed syntax-aware method tends to help select words near the leaves of dependency tree and avoids words near the root which are more likely important words in sentence. As our evaluations have shown that our proposed method is generally effective for performance improvement, which indicates that NMT indeed has a data augmentation preference of revising those marginal words with less importance. 
Such an observation can be mirrored by multiple empirical comparisons in our experimental results.

We assume that the automatically generated data by a data augmentation method may include both useful and harmful parts for machine learning. In the case of NMT, the harmful data will be likely from those sentences after revising key words in the original sentence. Note that among all three baseline data augmentation methods, the method \textit{Blanking} is better than any other, while \textit{Replacement} is the worst, between which there is a performance gap of 1.0 BLEU score. The method \textit{Blanking} gives the least change to a sentence among all the three methods, while 
the method \textit{Replacement} may put new word in the sentence, which has a chance to change the meaning of sentence completely and even hurt its syntactic structure. 
The method \textit{Dropout} takes a moderate revising over sentence by simply removing partial information of words, which will not alter the original sentence as violently as the method \textit{Replacement}. Thus, it is not a coincidence that the performance improvement from the three data augmentation methods has the same ranking list as the same as the degree of change to sentence by them. Overall, data augmentation for NMT prefer to those less syntactically and less semantically modified data.

Compared with others, the method \textit{Blank} keeps syntactic structure of the original sentence unchanged and does not introduce potentially harmful information as \textit{Replacement}. New words replaced in the method \textit{Replacement} may dramatically change the meaning of sentence which makes it perform quite unsatisfactorily. With our method to adjust the probabilities of selecting words, more than keeping those key words of sentence, our enhancement method can reduce the negative impacts of the original data augmentation methods in some way. 

\section{Conclusions}
In this work, we have presented a novel syntax-aware enhancement of robustness-oriented data augmentation for NMT, which is capable of setting sentence-specific probability on word selection 
for word-level revising. 
We take the depth of word in dependency parse tree to give the initial clue for word importance measuring, so that less important words to the its sentence may be more likely selected. 
Our proposed method is conceptually simple, easily implemented, and conveniently incorporated to standard word level data augmentation method. 
Our method is evaluated on WMT14 En-De dataset and IWSLT14 De-En dataset. To evaluate the impact of timing of data augmentation on performance, we do data augmentation before and during training.  The result of extensive experiments show our method can outperform strong baselines by effectively enhancing standard data augmentation methods. 

\bibliography{anthology,emnlp2020}
\bibliographystyle{acl_natbib}

\end{document}